\documentclass[journal]{IEEEtran}

\usepackage{graphicx}
\usepackage{amssymb,amsmath,epsfig}
\usepackage{algorithm}
\usepackage{algorithmic}
\usepackage{booktabs}
\usepackage{mathrsfs}
\usepackage{color}
\usepackage{textcomp}
\usepackage{bbding}
\usepackage{pifont}
\usepackage{wasysym}
\usepackage{amssymb}
\usepackage{subcaption}
\usepackage{cite}
\usepackage{amsmath,amssymb,amsfonts}
\usepackage{xcolor}
\usepackage{multirow}
\usepackage{epsfig}
\usepackage{array}
\usepackage{hyperref}
\begin{document}
%
\title{JDSR-GAN: Constructing An Efficient Joint Learning Network for Masked Face Super-Resolution }
%
%
%
\author{Guangwei Gao,~\IEEEmembership{Senior Member,~IEEE,}
        Lei Tang,
        Fei Wu,
        Huimin Lu,~\IEEEmembership{Senior Member,~IEEE}\\
        and Jian Yang,~\IEEEmembership{Member,~IEEE}

\thanks{This work was supported in part by the National Natural Science Foundation of China under Grant nos. 61972212 and 62076139, and Open Fund Project of Provincial Key Laboratory for Computer Information Processing Technology (Soochow University) (No. KJS2274). (\textit{Guangwei Gao and Lei Tang contributed equally to this work.) (Corresponding author: Fei Wu.)}}
\thanks{Guangwei Gao and Lei Tang are with the Institute of Advanced Technology, Nanjing University of Posts and Telecommunications, Nanjing, China, and also with the Provincial Key Laboratory for Computer Information Processing Technology, Soochow University, Suzhou, China (e-mail: csggao@gmail.com, tl\_njupt@163.com).}
\thanks{Fei Wu is with the College of Automation, Nanjing University of Posts and Telecommunications, Nanjing, China (e-mail: wufei\_8888@126.com).}
\thanks{Huimin Lu is with the Department of Mechanical and Control Engineering, Kyushu Institute of Technology, Kitakyushu 804-8550, Japan (e-mail: dr.huimin.lu@ieee.org).}
\thanks{Jian Yang is with the School of Computer Science and Technology, Nanjing University of Science and Technology, Nanjing, China (e-mail: csjyang@njust.edu.cn).}
}

\markboth{IEEE Transactions on Multimedia}%
{Shell \MakeLowercase{\textit{et al.}}: Bare Demo of IEEEtran.cls for IEEE Journals}
%

\maketitle

\begin{abstract}
With the growing importance of preventing the COVID-19 virus in cyber-manufacturing security, face images obtained in most video surveillance scenarios are usually low resolution together with mask occlusion. However, most of the previous face super-resolution solutions can not efficiently handle both tasks in one model. In this work, we consider both tasks simultaneously and construct an efficient joint learning network, called JDSR-GAN, for masked face super-resolution tasks. Given a low-quality face image with mask as input, the role of the generator composed of a denoising module and super-resolution module is to acquire a high-quality high-resolution face image. The discriminator utilizes some carefully designed loss functions to ensure the quality of the recovered face images. Moreover, we incorporate the identity information and attention mechanism into our network for feasible correlated feature expression and informative feature learning. By jointly performing denoising and face super-resolution, the two tasks can complement each other and attain promising performance. Extensive qualitative and quantitative results show the superiority of our proposed JDSR-GAN over some competitive methods.
\end{abstract}


%
\IEEEpeerreviewmaketitle

\section{Introduction}
\label{sec1}

Recently, most people are suffering from the outbreak of novel coronavirus 2019 (COVID-19). The world health organization (WHO) has pointed out that wearing a mask is an effective way to prevent the spread of the COVID-19 virus. With the improvement awareness of epidemic prevention, face images captured in conventional unlimited scenes such as video surveillance possess complex variations such as mask and low-resolution (LR) simultaneously. Obtaining high-resolution (HR) face images without the mask is now an essential yet challenging task, which plays an important role in many face-related security applications, e.g., face alignment~\cite{wan2020robust}, face parsing~\cite{GFC}, face detection~\cite{chaudhuri2019joint}, face tracking~\cite{zhu2020learning}, and face recognition~\cite{gao2017learning,gao2020hierarchical,zhao2021incremental}. Although many existing approaches have achieved promising progress in attaining high-quality HR face samples from the related low-quality LR ones~\cite{ma2020deep,chen2020learning,li2020learning,lu2021face,gao2022context,wang2022propagating}, most of them can only be used to handle one type of variation, such as LR face super-resolution or masked face image completion. In practice application scenarios (e.g., video surveillance), these approaches may not be applicable to the case where both LR and masked face are attained simultaneously. 

\begin{figure}[t]  
	\centerline{\includegraphics[width=8.8cm]{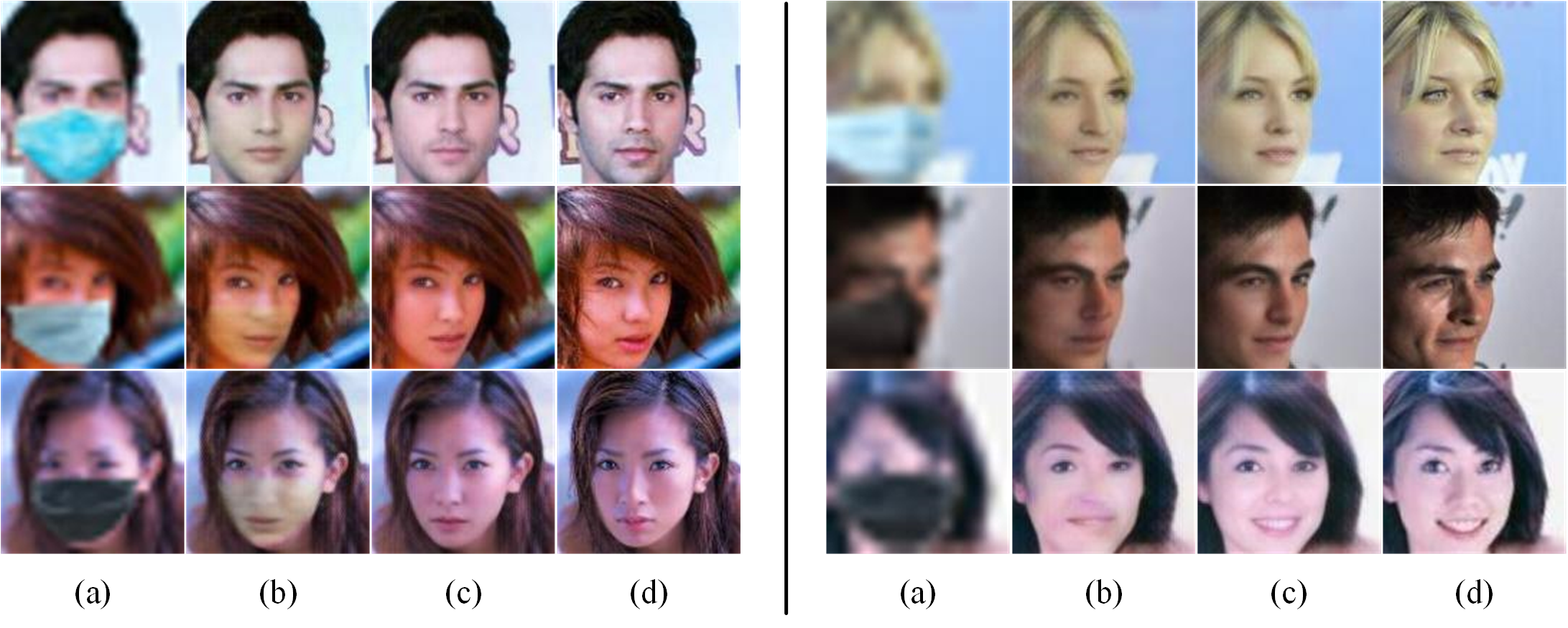}}
	\caption{Some results. In each panel, (a) is the input masked low-quality face images, (b) and (c) are the super-resolved images by applying denoising (CBDNet~\cite{CBD}) and face super-resolution (FSRNet~\cite{FSR}) successively and by our proposed JDSR-GAN, (d) is the high-resolution face images.}
	\label{fig1}
\end{figure}

One alternative way to deal with masked face super-resolution task is to perform image denoising followed by the face super-resolution procedure. However, it is not known whether the denoising methods are feasible for the LR face images. Meanwhile, the efficiency of existing face super-resolution solutions is not explicit when they are used to super-resolve LR face images with a mask. As shown in Fig.~\ref{fig1}, when a denoising algorithm (CBDNet~\cite{CBD}) and a face super-resolution algorithm (FSRNet~\cite{FSR}) are utilized in sequence to an observed masked LR face image, the super-resolved face images (Fig. 1 (b)) may miss some facial details to a certain extent. This straightforward recovering scheme maybe not optimal because it performs denoising and super-resolution separately, which may ignore the collaborative properties of these two tasks during the recovery procedure.

Different from these existing solutions, our target is to tackle a more challenging problem of how to super-resolve high-quality face images from both LR and masked face inputs in a single model. To this end, in this work, we design an end-to-end joint cooperation  framework via a generative adversarial network (GAN)~\cite{guan2019cooperation}. Through the generator, we can perform face image denoising and super-resolution simultaneously to obtain high-quality HR face images without mask from input masked low-quality face image. 
In summary, the main contributions of this work can be concluded in three-fold: 
\begin{itemize}
\item We introduce identity loss and attention mechanism into our denoising and super-resolution models. Thus, our designed network can refine faithful facial features and obtain better reconstruction performance. 
\item We devise an effective framework for jointly performing denoising and face super-resolution via a single model. Thus the two parts can provide collaborative and complementary information to each other for better restoration. 
\item We obtain promising masked face super-resolution results compared with some existing face super-resolution approaches especially for the low-quality face images obtained from real-world scenes.
\end{itemize}

\section{Related work}
\label{sec2}


\subsection{Image Denoising}
\label{sec21}

Recently, on account of the remarkable achievement of deep neural networks in image classification, image denoising approaches based on deep learning have been well developed~\cite{Liao_2021_CVPR}. Zhang et al.~\cite{DNCNN} combined residual learning~\cite{Residual} and batch normalization~\cite{BN} to propose a denoising model addressing the gradient dispersion caused by deepening of the network layers. Furthermore, the noise in practical images is derived from various scenes. Blind denoising of practical noisy images is still a challenging task. 
Zhu et al.~\cite{2016From} proposed to model image noise using a mixed Gaussian (MoG) model and developed a low-rank MoG filter to recover clean images. 

Zhang et al.~\cite{CBD} proposed a CBDNet composed of a noise estimation sub-net and a non-blind denoising sub-net, where the asymmetric loss was introduced to suppress underestimation errors of noise levels. In addition to noise simulation of RGB images, Brooks et al.~\cite{2019unprocessing} analyzed the image signal processing channel and then generated raw images directly by inverting each step of an image processing pipeline. Tian et al.~\cite{2020image} exploited residual learning, dilated convolutions, and batch re-normalization to tackle the real noisy image. Wang et al.~\cite{2020practical} proposed a novel k-Sigma transform that allows the model to remove the ISO constraint, enabling the small network to efficiently tackle an extensive range of noise levels.

\subsection{Image Super-Resolution}
\label{sec22}

The target of the single image super-resolution (SR) is to recover HR images from corresponding LR inputs. In recent years, deep neural networks have been broadly adopted for the super-resolution task. Ledig et al.~\cite{2017Photo} presented a generative adversarial network based method for photo-realistic images super-resolution by utilizing a perceptual loss function. Li et al.~\cite{li2019feedback} designed an image super-resolution feedback network (SRFBN) to achieve a better SR performance. Guo et al.~\cite{guo2020closed} proposed a dual regression network (DRN) by introducing an additional dual regression mapping on LR data. Gao et al.~\cite{gao2022feature} proposed a lightweight feature distillation interaction weighted network for efficient image SR tasks, striking a good balance between model performance and efficiency. 

Face image super-resolution is a class-specific image recognition method that exploits the statistical properties of face images~\cite{gao2020constructing,liu2022hallucinating,gao2022ctcnet}. Earlier techniques assumed that faces are in a controlled environment with tiny variations. Yu et al.~\cite{2018Super} embedded attributes in the procedure of face image super-resolution. Chen et al.~\cite{FSR} and Song et al.~\cite{2017Learning} both used a multi-task approach for coarse-to-fine face super-resolution. Then, Zhang et al.~\cite{SICNN} introduced a super identity loss to evaluate the differences of identity information. Hsu et al.~\cite{hsu2019sigan} leveraged the facial identity information for identity-preserving face SR task. Recently, Ma et al.~\cite{ma2020deep} propose a  face SR method with iterative collaboration between facial image recovery and landmark estimation.


\begin{figure}[!t]
	\centerline{\includegraphics[width=8.6cm]{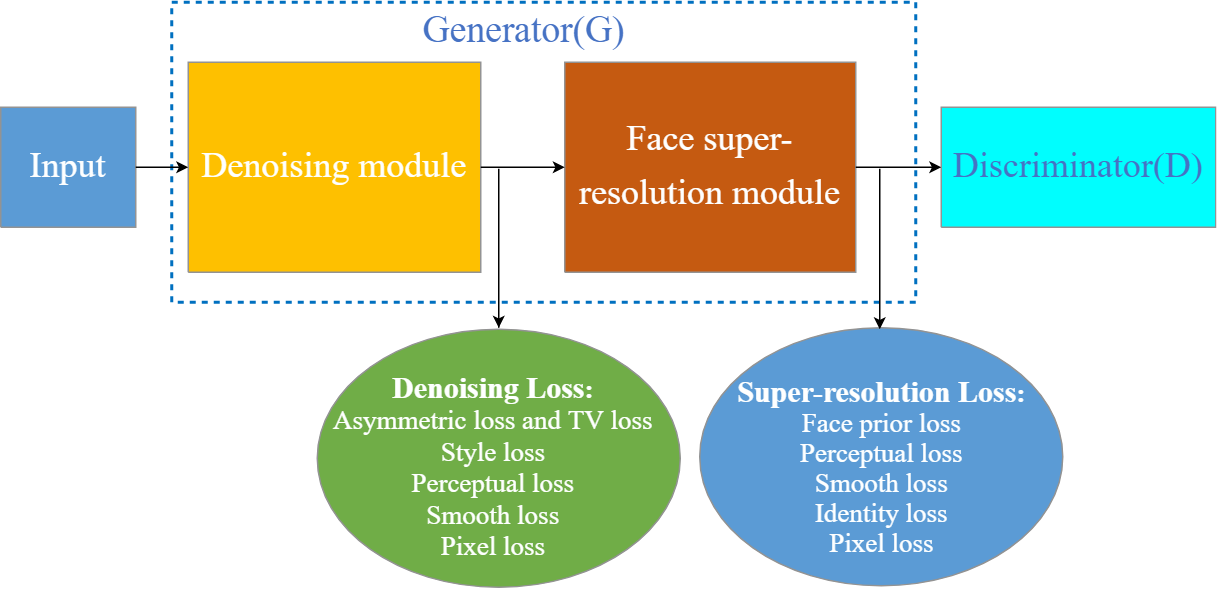}}
	\caption{Network structure of our JDSR-GAN method. The whole network is jointly trained end-to-end by collaboratively using denoising loss, super-resolution loss and adversarial loss.}
	\label{fig2}
\end{figure}

\section{Proposed Method}
\label{sec3}

Fig.~\ref{fig2} depicts the whole pipeline of our proposed method, which is composed of a generator, a discriminator, and the related losses. 

\subsection{Network Architecture}
\label{sec31}

\begin{figure}[!t]
	\centerline{\includegraphics[width=8.7cm]{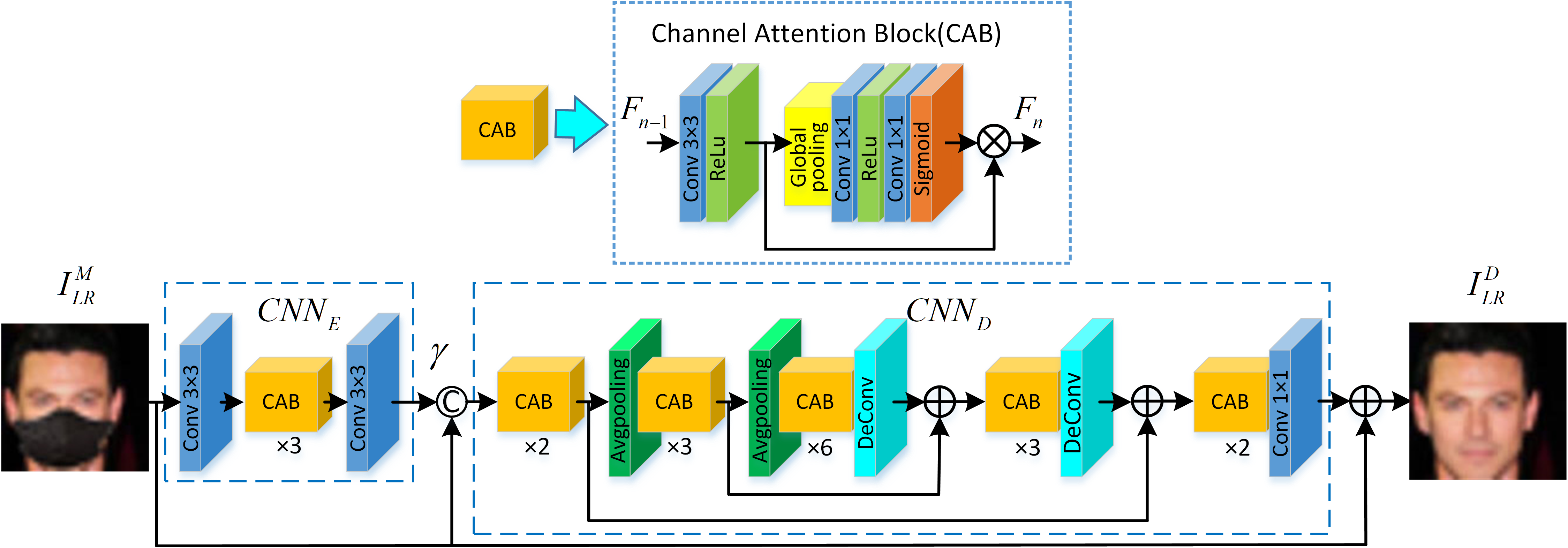}}
	\rule[1pt]{8.8cm}{0.05em}
	\centerline{\includegraphics[width=8.7cm]{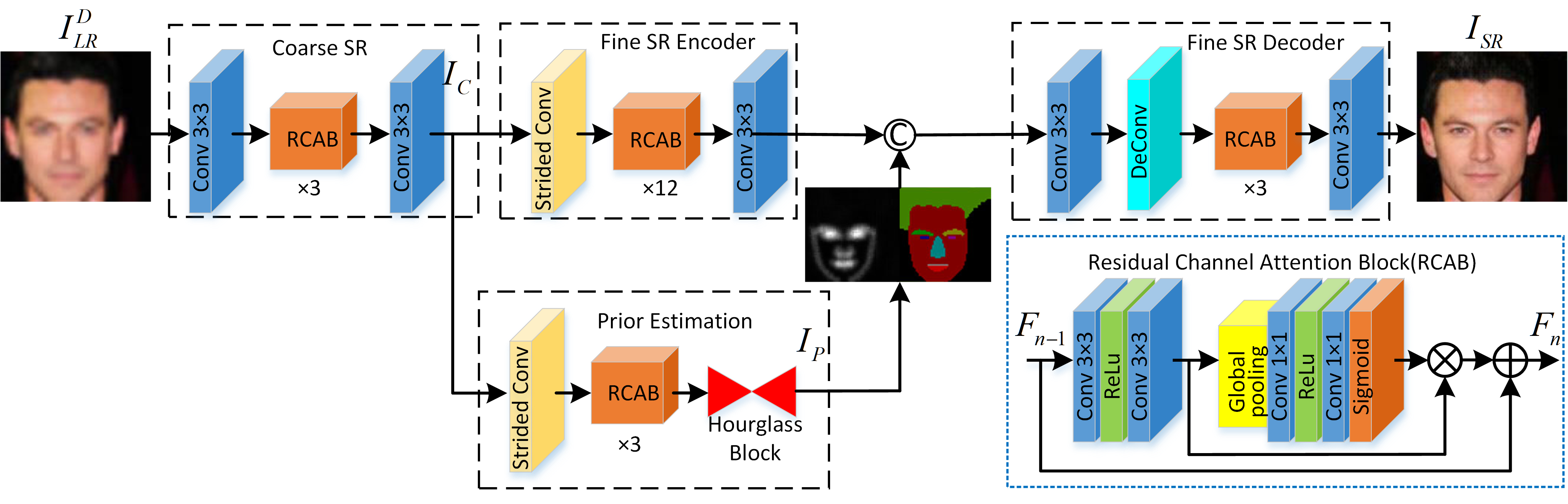}}
	\caption{Network structure of the denoising module (top) and face super-resolution module (bottom). ``Conv'' embedded in the main stream depicts a convolutional layer together with the Batch Normalization~\cite{BN} and ReLU~\cite{ReLU} operations. ``Strided Conv'' indicates the convolutional layer with the size of the kernel be 3×3 and the stride to be 2. }
	\label{fig4}
\end{figure}

\textbf{Denoising Module:}
CBDNet~\cite{CBD} has achieved good performance at removing Gaussian noise but has not been studied for removing the mask in face images.	The channel attention mechanism can be utilized to filter out the important points from a mass of information and enhance the capabilities of the network to identify different contributions of the feature maps. Based on the CBDNet, we add channel attention to each convolution block in the network to construct our denoising module. As illustrated in Fig.~\ref{fig4}, the denoising network can be decomposed into a noise evaluation subnetwork $CN{N_E}$ and a non-blind denoising subnetwork $CN{N_D}$, aiming to generate an LR non-masked image $I_{LR}^D$ from an input masked LR face image $I_{LR}^M$. The LR face image without mask addressed by the denoising module can be represented as 

\begin{equation}
\gamma {\rm{ = }}CN{N_E}(I_{LR}^M),
\label{eq1}
\end{equation}
\begin{equation}
I_{LR}^D = CN{N_D}([\gamma ,I_{LR}^M]) + I_{LR}^M,
\label{eq2}
\end{equation}
where [·], and $\gamma $ denote the procedure of concatenation and estimated noise level map respectively.

\textbf{Face Super-Resolution Module:}
After the denoising module, the face image $I_{LR}^D$ is fed into the following super-resolution module to get a high-quality face image without the mask. Similar to the previous operations, we introduce channel attention into each residual block as show in Fig.~\ref{fig4}. The face super-resolution module is composed of a coarse-SR network, a prior estimation network, an encoder, and a decoder network, which takes the geometry prior, i.e., face parsing maps and facial landmark heatmaps into consideration. The process of face super-resolution can be formulated as

\begin{equation}
{I_C} = Coarse(I_{LR}^D),
\label{eq3}
\end{equation}
\begin{equation}
{I_P} = Prior({I_C}),
\label{eq4}
\end{equation}
\begin{equation}
{I_{Mix}} = [Encoder({I_C}),{I_P}],
\label{eq5}
\end{equation}
\begin{equation}
{I_{SR}} = Decoder({I_{Mix}}),
\label{eq6}
\end{equation}
where ${I_C},{I_P},{I_{Mix}}$, and ${I_{SR}}$ represents the coarse SR image recovered from $I_{LR}^D$, prior estimation evaluated from ${I_C}$, the concatenation of image feature and prior estimation, and the final output high-resolution non-masked face image.

\textbf{Generator and Discriminator:}
Images generated by conventional super-resolution methods lack high-frequency information and fine details, which can only be remedied by selecting the appropriate target functions. While GAN can solve this problem, it has exhibited great potential in super-resolution, generating photo-realistic images with superior visual effects~\cite{2017Photo}. As depicted in Fig.~\ref{fig2}, the generator of our JDSR-GAN consists of an image denoising module and successively a super-resolution module. Ideally, given an observed low-quality masked face image $I_{LR}^M$, the output face image by the generator should be a non-masked face image with high resolution. 

We use a discriminator network to distinguish the real HR images and the super-resolved ones, which plays an auxiliary character in our network training. The structure of our discriminator is the same as that in WGAN-GP~\cite{WGAN-GP}. WGAN-GP removes weight clipping from WGAN~\cite{WGAN} and adds the gradient penalty to discriminator loss, enabling the networks to converge fast and stably. The loss function of our discriminator is given as

\begin{equation}
\begin{array}{l}
L_{adv}^{HR} = \mathop {\min }\limits_G \mathop {\max }\limits_D  - {{\rm E}_{{x_r}\sim{p_r}}}[D({x_r})] + {{\rm E}_{{x_g}\sim{p_g}}}[D({x_g})]\\
			 ~~~~~~+ \eta {{\rm E}_{\hat x\sim{p_{\hat x}} }}[{(||{\nabla _{\hat x}}D(\hat x)|{|_2} - 1)^2}],
\end{array}
\label{eq7}
\end{equation}
where $D$ and $G$ represent the discriminator and generator respectively. ${p_r}$ denotes the distribution of the real face images, ${p_g}$ denotes the generator distribution implicitly defined by ${x_g} = G(z)$, $z\sim{p_z}$ (${p_z}$ denotes the distribution of the masked face images) and ${p_{\hat x}}$ can be defined as the data distribution sampled from ${p_r}$ and ${p_g}$. ${\nabla _{\hat x}}$ denotes the gradient operator. $\eta$ denotes the penalty coefficient, which is set as 0.1. 

In our experiments, extensive evaluations have proven that our proposed approach is feasible and effective. Our multi-task training strategies take advantage of the complementary information of the two tasks so that we can obtain fine-grained face recovery images with fewer artifacts. Moreover, we also need to carefully design appropriate loss functions for the entire network. We will detail these in the next part.
	
\subsection{Loss Functions}
\label{sec32}

\textbf{Asymmetric loss and total variation (TV) regularization.}
The non-blind denoising model is very sensitive to noise level, so we introduce asymmetrical loss into the noise estimation subnetwork to avoid estimation error of noise level. 
The asymmetric loss is defined as 

\begin{equation}
L_{LR}^{asym} = \sum\limits_{i = 0}^{N - 1} {|\alpha  - {\mathbb{I} _{(\hat \gamma (y_i) - \gamma (y_i))}}|}  \cdot {(\hat \gamma (y_i) - \gamma (y_i))^2},
\label{eq8}
\end{equation}
where ${\mathbb{I} _{(\hat \gamma (y_i) - \gamma (y_i))}}$ represents a mathematical expression when $\mathbb{I}  = 1$ for $\hat \gamma (y_i) - \gamma (y_i) < 0$ and 0 otherwise, ${\hat \gamma (y_i)}$, ${\gamma (y_i)}$ represent the estimated noise level and corresponding ground truth at pixel $i$ respectively, $\rm{y}$ represents the synthetic noisy image, and $\alpha$ is a parameter set between 0 and 0.5.

Since many recovery algorithms amplify the noise, we incorporate a total variation regularization, which constrains the smoothness of the image pixels to ensure that the horizontal and vertical pixel changes of the image shrink to a certain range. The TV loss can be defined as 

\begin{equation}
L_{LR}^{TV}{\rm{ = ||}}{\nabla _h}\hat \gamma {\rm{(y)||}}_2^2 + {\rm{||}}{\nabla _v}\hat \gamma {\rm{(y)||}}_2^2,
\label{eq9}
\end{equation}
where ${\nabla _v}$ and ${\nabla _h}$ represent the gradient operator along the vertical direction and horizontal direction respectively.

\textbf{Pixel loss.} In fact, ${L_2}$ loss posses a strong penalty for large errors and a weak penalty for small errors, neglecting the impact of the image content itself, i.e., generates smoother images. However, when distinct textures appear, then the result of optimizing ${L_2}$ loss can easily blur this area. Furthermore, the convergence performance of ${L_2}$ loss is worse than that of ${L_1}$ loss. Thus, the pixel loss can be defined as 

\begin{equation}
\left\{ {\begin{array}{*{20}{c}}
{L_{pixel}^{LR} = ||I_{LR}^{GT} - I_{LR}^D|{|_1}}\\ \\
{L_{pixel}^{HR} = ||I_{HR}^{GT} - {I_{SR}}|{|_1}},
\end{array}} \right.
\label{eq10}
\end{equation}
where $|| \cdot |{|_1}$ denotes the ${L_1}$ norm, $I_{LR}^{GT}$ and $I_{HR}^{GT}$ denote the ground-truth non-masked LR face image and the ground-truth non-masked HR face image respectively. 

\textbf{Perceptual loss.} Previous super-resolution methods mostly used mean square error (MSE) as loss function. Although good super-resolution results can be obtained by minimizing MSE loss, it may be difficult to avoid fuzzy details, which is caused by the flaws of MSE itself. Thus, we use perceptual loss here, which will make the restored image look better in visual effect. The perceptual loss is formulated as
\begin{equation}
\left\{ {\begin{array}{*{20}{c}}
{L_{per}^{LR} = \frac{1}{{{W_{i,j}}{H_{i,j}}}}\sum\limits_{n = 1}^N {||{\phi _{i,j}}(I_{LR}^{GT}) - {\phi _{i,j}}(I_{LR}^D)|{|_1}} }\\ \\
{L_{per}^{HR} = \frac{1}{{{W_{i,j}}{H_{i,j}}}}\sum\limits_{n = 1}^N {||{\phi _{i,j}}(I_{HR}^{GT}) - {\phi _{i,j}}({I_{SR}})|{|_1}} },
\end{array}} \right.
\label{eq11}
\end{equation}
where $\phi$ denotes ${\rm{VGG16}}$~\cite{VGG-16} pre-trained on ImageNet~\cite{2015ImageNet}, ${\phi _{i,j}}$ denotes the feature from the ${j-\rm{th}}$ convolution layer ahead of the ${i-\rm{th}}$ max pooling layer, ${W_{i,j}}$ and ${H_{i,j}}$ denote the size of the map mentioned above.

\textbf{Smooth loss.} When we conduct face image denoising, the obtained images may exhibit trivial color distortions around the boundaries of the masked area. Thus, we also incorporate the smooth loss to alleviate such distortions. 
The formula is as follows

\begin{equation}
\left\{ {\begin{array}{*{20}{c}}
{\begin{array}{*{20}{c}}
{L_{smooth}^{LR} = \sum\limits_{i = 0}^W {\sum\limits_{j = 0}^H {||\;I_{LR}^D(i,j + 1) - I_{LR}^D(i,j)|{|_1}} } }\\
~~~~~~~~~~~{ + \sum\limits_{i = 0}^W {\sum\limits_{j = 0}^H {||\;I_{LR}^D(i + 1,j) - I_{LR}^D(i,j)|{|_1}} } }
\end{array}}\\ \\
{\begin{array}{*{20}{c}}
{L_{smooth}^{HR} = \sum\limits_{i = 0}^W {\sum\limits_{j = 0}^H {||\;{I_{HR}}(i,j + 1) - {I_{HR}}(i,j)|{|_1}} } }\\
~~~~~~~~~~~{ + \sum\limits_{i = 0}^W {\sum\limits_{j = 0}^H {||\;{I_{HR}}(i + 1,j) - {I_{HR}}(i,j)|{|_1}} } },
\end{array}}
\end{array}} \right.
\label{eq12}
\end{equation}
where $H$ and $W$denote the height and width of the recovered image, respectively.

\textbf{Style loss.} During the process of denoising, an essential task is to render the style of the denoising area that looks similar enough to the non-masked area. Thus, we incorporate the style loss~\cite{Style} into the denoising module which works by merging the contextual content of the output image with that of the ground-truth one. 
The style loss is defined as

\begin{small}
\begin{equation}
\begin{array}{l}
L_{style}^{LR} = \sum\limits_{n = 1}^N {||{F_n}({\phi _n}{{(I_{LR}^{GT})}^T}{\phi _n}(I_{LR}^{GT}) - {\phi _n}{{(I_{LR}^D)}^T}{\phi _n}(I_{LR}^D))|{|_1}} ,
\end{array}
\label{eq13}
\end{equation}
\end{small}
where ${F_n}$ is a normalization factor $1/({C_n} \cdot {W_n} \cdot {H_n})$ for the ${n-{th}}$ ${\rm{VGG16}}$ layer. ${C_n}$, ${W_n}$ and ${H_n}$ denote the channel number, width and height of the maps, respectively. 

\textbf{Face prior loss.} 
The network introduces two related face priors, face parsing and face landmark, as the supplementary evaluation metrics, penalizing the discrepancy between the geometry of the generated images and the ground-truth ones. The named face prior loss is formulated as

\begin{equation}
\begin{array}{r}
L_{fp}^{HR} = \mu ||{L_{landmark\_p}} - {L_{landmark\_gt}}|{|_2}\\
 + \nu ||{H_{parsing\_p}} - {H_{parsing\_gt}}|{|_2},
\end{array}
\label{eq14}
\end{equation}
where ${H_{parsing\_p}}$, ${L_{landmark\_p}}$, ${H_{parsing\_gt}}$ and ${L_{landmark\_gt}}$  denote the estimated face parsing maps and face landmark maps from the recovered images, the referenced face parsing maps and face landmark heatmaps, respectively. Empirically, we set $\mu  = 1$ and $\nu  = 0.1$.

\textbf{Identity loss.} Pioneer work~\cite{SICNN} has revealed that identity is an important criterion in terms of distinguishing each object. We expect that the super-resolved images have a similar identity as their target ones. Thus, we further introduce identity loss into the training process, aiming to enhance image fidelity and identity recognition. In this paper, we use a Resnet-like $CNN$~\cite{Residual} as the face feature extraction network (denoted as $CNN_E$). The identity loss can be defined as

\begin{equation}
L_{identity}^{HR} = ||CN{N_E}({I_{SR}}) - CN{N_E}(I_{HR}^{GT})|{|_2},
\label{eq15}
\end{equation}
where $CN{N_E}({I_{SR}})$ and $CN{N_E}(I_{HR}^{GT})$ are the identity features of images ${I_{SR}}$ and $I_{HR}^{GT}$ extracted by the model $CN{N_E}$.

\subsection{Training Strategy}
\label{sec33}

As shown in Fig.~\ref{fig2}, we devise a multi-task training network. The denoising module integrates the asymmetric loss, TV loss, style loss, pixel loss, perceptual loss, and smooth loss. The entire loss function at this stage can be represented as 

\begin{equation}
\begin{array}{c}
\begin{array}{c}
{L_{de}} = \lambda _1^1L_{asym}^{LR} + \lambda _1^2L_{TV}^{LR} + \lambda _1^3L_{style}^{LR} + \lambda _1^4L_{per}^{LR}\\
 + \lambda _1^5L_{pixel}^{LR} + \lambda _1^6L_{smooth}^{LR},
\end{array}
\end{array}
\label{eq16}
\end{equation}
where $\lambda _1^1$, $\lambda _1^2$, $\lambda _1^3$, $\lambda _1^4$, $\lambda _1^5$ and $\lambda _1^6$ represent the weight of individual losses. For asymmetric loss and TV loss, we follow~\cite{CBD} and set $\lambda _1^1 = 0.5$ and $\lambda _1^2 = 0.05$.  For other losses, we set $\lambda _1^3 = 10$, $\lambda _1^4 = 0.1$, $\lambda _1^5 = 1$, and $\lambda _1^6 = 1$.

For the face image super-resolution module, we apply some losses from the previous module, such as style loss, pixel loss, perceptual loss, and smooth loss. Furthermore, we add face prior loss, identity loss, adversarial loss, and the entire loss can be denoted as 

\begin{equation}
\begin{array}{c}
{L_{fsr}} = \lambda _2^1L_{fp}^{HR} + \lambda _2^2L_{per}^{HR} + \lambda _2^3L_{pixel}^{HR} + \lambda _2^4L_{smooth}^{HR}\\
 + \lambda _2^5L_{identity}^{HR} + \lambda _2^6L_{adv}^{HR},
\end{array}
\label{eq17}
\end{equation}
where $\lambda _2^1$, $\lambda _2^2$, $\lambda _2^3$, $\lambda _2^4$, $\lambda _2^5$ and $\lambda _2^6$ denote the weight of different losses. For perceptual loss and the smooth loss, we also follow~\cite{2017Photo} and empirically set $\lambda _2^2 = 0.1$, $\lambda _2^4 = 0.01$. For  face prior loss and pixel loss, we also follow~\cite{FSR} and set $\lambda _2^1 = 1$ and $\lambda _2^3 =1$. For other losses, we set $\lambda _2^5 =1$ and $\lambda _2^6 = {10^{ - 3}}$.

For the entire network, ${L_{de}}$ and ${L_{fsr}}$ are integrated to make the denoising module and face super-resolution complement each other. The total loss can be represented as 

\begin{equation}
{L_{total}} = {L_{de}} + {L_{fsr}}.
\label{eq18}
\end{equation}

\begin{figure}[t]  
	\centerline{\includegraphics[width=8.7cm]{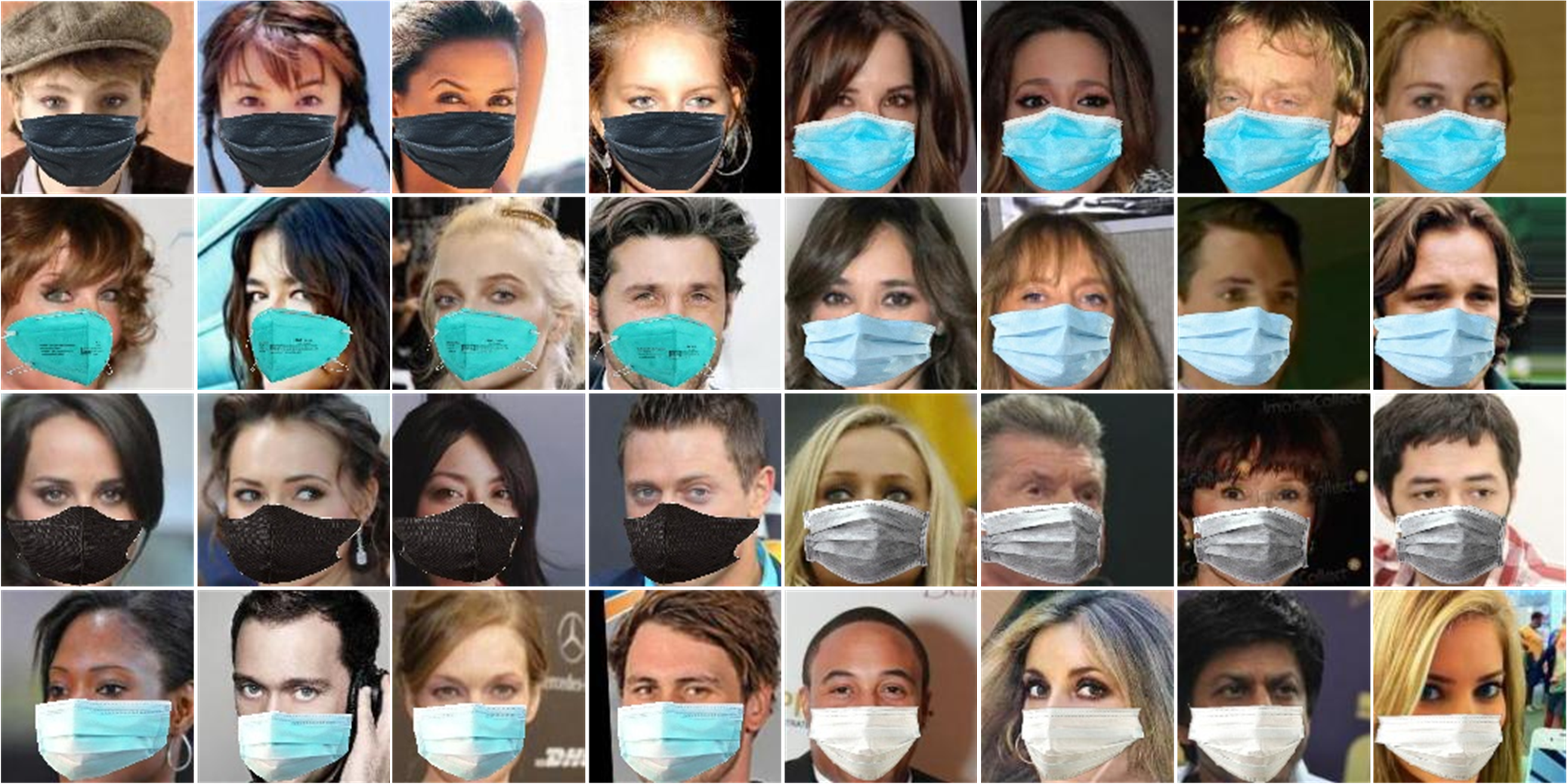}}
	\caption{ Some artificially masked training examples in the CelebA dataset.}
	\label{fig5}
\end{figure}

\section{Experimental Evaluations}
\label{sec4}

\subsection{Dataset and Metrics}
\label{sec41}

We validate the performance of respective methods on CelebA~\cite{CelebA} face dataset. CelebA is a widely used large-scale dataset that contains 10,177 face objects and 202,599 samples. Following the previous standard protocol, we use 162,770 to construct the training set, 19,867 images to construct the validation set, and 19,962 images test set. In real-world application scenes, it is unreasonable to acquire coupled face images, i.e., clean face samples and their corresponding faces with the mask. To obtain the faces with the mask, we first use a face detection method~\cite{Dlib} to detect the location of key points and perform face alignment operation in each face of CelebA, and then calculate the position of the mask in the face based on the coordinates of the nose, left and right cheeks and jaw. Finally, we scale the mask image to an appropriate size to fuse with the face image. Some examples of masked faces are given in Fig.~\ref{fig5}. The similarity between the ground-truth face images and recovered ones are evaluated in terms of SSIM and PSNR~\cite{SSIM}, which are evaluated on the Y channel in the converted YCbCr space. We also give the FID index~\cite{heusel2017gans} to evaluate the visual quality of the face images.

\begin{table*}[t]
	\caption{ Ablation study of different modules.}
	\begin{center}
		\setlength\tabcolsep{6pt} 
		{\fontsize{9pt}{16pt}\selectfont
		\begin{tabular}{ >{\centering}p{1.5cm} >{\centering}p{2cm} >{\centering}p{2cm} >{\centering}p{2cm} >{\centering}p{2cm} >{\centering}p{2cm} p{2cm}<{\centering} }
         \hline
			Model  & W/o ${L_{style}}$ & W/o ${L_{per}}$ & W/o ${L_{identity}}$ & W/o ${L_{smooth}}$ & W/o attention & JDSR-GAN \\ \hline \hline
			PSNR (dB)  &25.85  &25.86   &26.22  &26.19  &26.19   &26.28\\
			SSIM       &0.8104 &0.8119   &0.8118 &0.8076 &0.8109  &0.8134\\\hline

		\end{tabular}}
		\label{tab2}
	\end{center}
\end{table*}

\subsection{Implementation Details}
\label{sec42}

To obtain the ground truth of face parsing maps on CelebA dataset, we utilize GFC~\cite{GFC} trained on the Helen~\cite{Helen} dataset as the face parsing instrument to estimate the parsing results. During the pre-training of the face parsing network, we explore Adam~\cite{2014Adam} method with an initial learning rate as $10^{- 4}$. For the ground truth of facial landmarks on CelebA, we also exploit the public available SeetaFace model to estimate the 81 landmarks for each face image. For the multi-task training, we crop and normalize the face regions in CelebA dataset to the size $128 \times 128$. Then we add a mask into each face image and downsample these masked face images into the size of $32 \times 32$ (4 times) or $16 \times 16$ (8 times) as the degraded inputs. 
Our experiments are developed based on Pytorch~\cite{Pytorch} using NVIDIA RTX 3090 GPUs.

\begin{table*}[t]
	\caption{ The objective indexes of respective methods on CelebA dataset. Red/blue indicates the best/second-best results.}
	\begin{center}
		\setlength\tabcolsep{15pt} 
		\renewcommand{\arraystretch}{1}
		{\fontsize{9pt}{15pt}\selectfont
		\begin{tabular}{ >{\centering}p{2cm} >{\centering}p{1cm} >{\centering}p{1.5cm} >{\centering}p{0.5cm} >{\centering}p{1cm} >{\centering}p{1cm}  p{2cm}<{\centering} }
       \hline
			Methods        & Factor &PSNR(dB)$\uparrow$  & SSIM$\uparrow$  &FID$\downarrow$  &Params$\downarrow$  &Multi-adds$\downarrow$  \\ \hline \hline 
            CBD+DRN     & $\times 4$  &26.48 &0.7398 &-     &14.4M &\textcolor{red}{34.90G}\\ 
            CBD+SRFBN   & $\times 4$  &26.59 &0.7443 &-     &\textcolor{blue}{8.0M}  &142.7G \\
            CBD+SICNN   & $\times 4$  &27.07 &0.7839 &-     &\textcolor{red}{7.5M}  &147.9G \\
            CBD+FSRGAN  & $\times 4$  &27.73 &0.8318 &11.34 &36.5M &52.30G \\
            CBD+DICGAN  & $\times 4$  &\textcolor{blue}{28.28} &\textcolor{blue}{0.8338} &\textcolor{blue}{8.33} &22.9M &155.9G \\
        \textbf{JDSR-GAN} & $\times 4$  &\textcolor{red}{29.18} &\textcolor{red}{0.8553} &\textcolor{red}{5.74} &36.7M &\textcolor{blue}{51.50G}
            \\\hline
            CBD+DRN     & $\times 8$  &23.61 &0.6371 &-     &\textcolor{red}{14.4M} &\textcolor{red}{20.20G} \\
            CBD+FSRGAN  & $\times 8$  &24.96 &0.7423 &38.13 &36.5M &52.30G \\
            CBD+DICGAN  & $\times 8$  &\textcolor{blue}{25.36} &\textcolor{blue}{0.7137} &\textcolor{blue}{33.71} &\textcolor{blue}{22.9M} &155.9G \\
      \textbf{JDSR-GAN} & $\times 8$  &\textcolor{red}{26.45} &\textcolor{red}{0.7633} &\textcolor{red}{17.68} &36.7M &\textcolor{blue}{51.50G} 
            \\\hline

		\end{tabular}}
		\label{tab3}
	\end{center}
\end{table*}

\begin{figure}[t]  
	\centerline{\includegraphics[width=8.7cm]{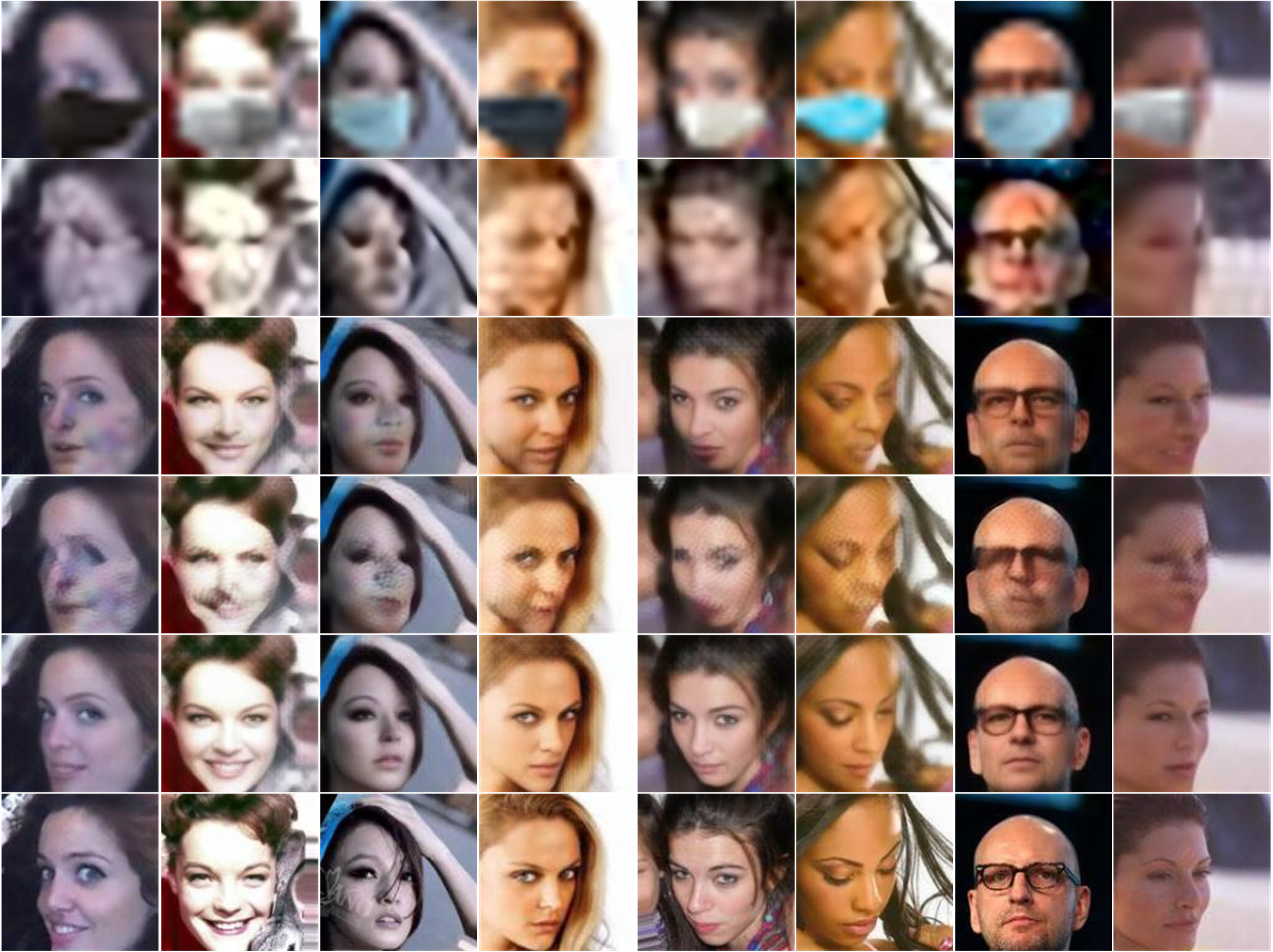}}
	\caption{ The qualitative comparisons between results obtained by respective methods on CelebA dataset with scale factor 8. From top to bottom are successively the input masked LR faces, the SR results of DRN~\cite{guo2020closed}, FSRGAN~\cite{FSR}, DICGAN~\cite{ma2020deep}, our JDSR-GAN and the ground-truth HR references. Better zoom in to see more details.}
	\label{fig8}
\end{figure}

\subsection{Ablation Study}
\label{sec43}

In our method, we have several loss functions and channel attention mechanism compared with previous related methods. In this part, we perform ablation experiments to assess the effectiveness of each component. All the studies are performed based on the same subset of the large-scale CelebA dataset, using the same masked low-quality face images with scale factor 4 (i.e., the size of the input is $32 \times 32$). The quantitative performance is tabulated in Table~\ref{tab2}. We can observe that when the model loses the constraint provided by the style loss and perceptual loss, the quality of the SR images is degraded since its ability to measure the reconstruction difference is weakened. A large improvement can also be observed from the channel attention, smooth loss, and identity information, which enables the network to flexibly capture the relationship between global and local features. The above ablation studies prove that each part of JDSR-GAN has an indispensable contribution to the improvement of the performance.

\subsection{Experimental Comparisons}
\label{sec44}

In this part, we compare our method with some state-of-the-art ones. The compared methods include two general image SR methods (SRFBN~\cite{li2019feedback} and DRN~\cite{guo2020closed}) and three face image SR methods (SICNN~\cite{SICNN}, FSRGAN~\cite{FSR}, and DICGAN~\cite{ma2020deep}). It is worthy that for those prominent SR methods, we first perform face denoising process on the LR inputs by the CBDNet~\cite{CBD} method. For a fair comparison, the CBDNet and those successive SR methods are pre-trained based on the same training set.


The qualitative comparisons of respective methods are listed in Fig.~\ref{fig8}. By considering the denoising and SR procedure separately, the results obtained by the compared methods have distinct noises in the masked area. In comparison, by integrating channel attention mechanism and some carefully designed losses (such as identity loss, face prior loss, style loss and perceptual loss), the SR images generated by our proposed JDSR-GAN can obtain quite better visual effects and can recover more facial details, especially for very low-quality face images (e.g., with the scale factor 8). Although the super-resolved face image is slightly different from the ground-truth ones around the mouth, the facial detail features are generally more similar to the ground-truth references. The quantitative comparisons are also given in Table~\ref{tab3}. By jointly performing denoising and SR task, our JDSR-GAN can attain remarkable PSNR, SSIM, and FID values than other compared methods, which further validate the superiority of our method. Also, we can observe that our JDSR-GAN can reach a good trade-off between accuracy and model size.


\subsection{Generality Study}
\label{sec45}

\begin{figure}[!t]  
	\centerline{\includegraphics[width=8.7cm]{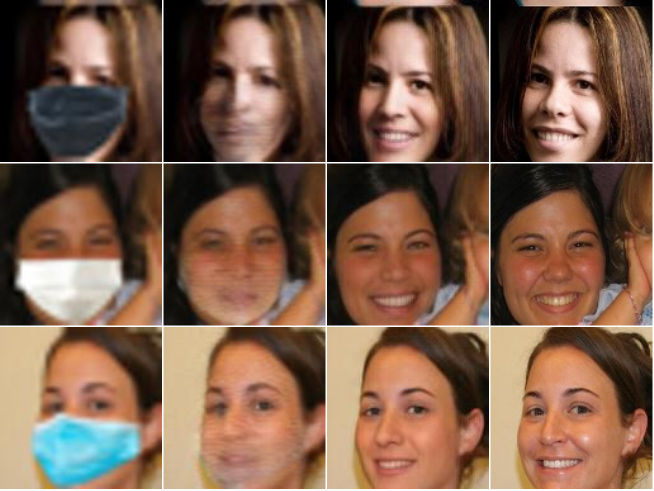}}
	\caption{ The visual comparisons between results obtained by JDSR-GAN and DICGAN on Helen dataset. For each person, from left to right are successively the input masked LR faces, the SR results of DICGAN~\cite{ma2020deep}, our JDSR-GAN and the ground-truth. Better zoom in to see more details.}
	\label{fig9}
\end{figure}

 In this part, we conduct experiments to study the generality of our JDSR-GAN. We use the model trained on CelebA to perform testing on Helen~\cite{Helen}. The masked face images have a size of $32 \times 32$. The visual comparisons of our JDSR-GAN and DICGAN are shown in Fig.~\ref{fig9}, from which we can observe that our method can attain more facial texture details around the masked area than the competitive ones, which generate many artifacts around the mouth. Specifically, the recovered faces by our JDSR-GAN look more similar to the ground-truth ones. In terms of the quantitative results, our JDSR-GAN achieves 25.5427dB PSNR, which is 1.6 dB higher than that of the DICGAN method.

\subsection{Results on Real-World Images}
\label{sec46}

\begin{figure}[t]  
	\centerline{\includegraphics[width=8.7cm]{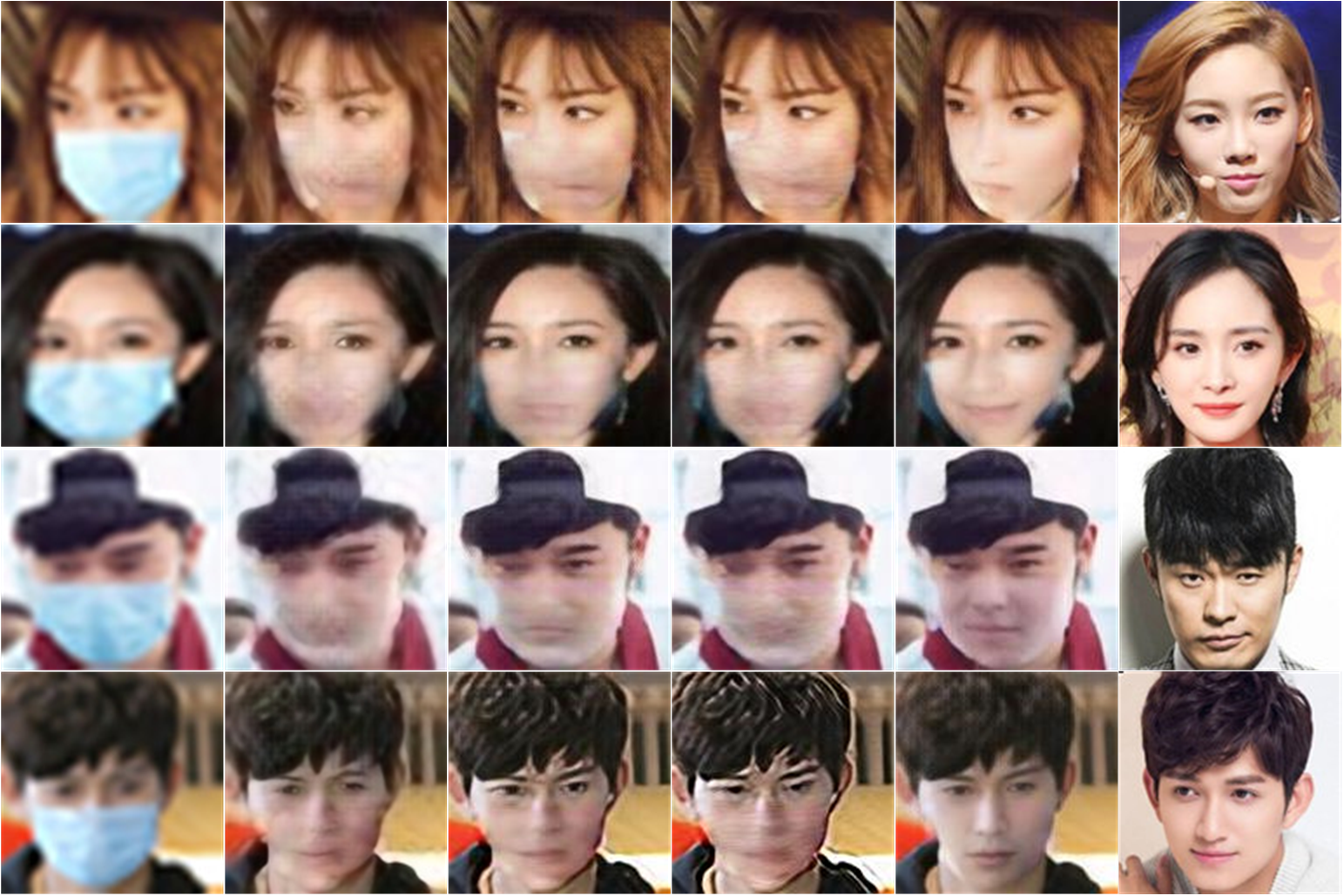}}
	\caption{ The visual comparisons of results obtained by respective methods on real low-quality images. For each person, from left to right are successively the input masked LR faces, the SR results of SICNN~\cite{SICNN}, FSRGAN~\cite{FSR}, DICGAN~\cite{ma2020deep}, our JDSR-GAN and the ``ground truth''.}
	\label{fig10}
\end{figure}

In all the above experiments, the masks in the faces are artificially added. In real application conditions, it is unreasonable and difficult for us to simulate the process of image degradation and wearing a mask. Thus, in this part, we perform experiments on real-world masked low-quality face images. The low-quality images are crawled from the Internet and resized to have a size of $128 \times 128$ as the inputs. The images of the same subject without a mask are regarded as the ``ground truth''. Fig.~\ref{fig10} shows the visual results of respective methods on several real low-quality images. Compared with other methods, our JDSR-GAN can obtain the best visual performance. It removes most of the mask and to some extent looks more similar to the ``ground truth''.

\section{Conclusions}

For the masked face super-resolution task, in this paper, we construct a joint learning network (named JDSR-GAN) to perform face image denoising and super-resolution simultaneously in a single model. Our JDSR-GAN method uses multi-task learning to integrate the channel attention mechanism and some carefully designed losses to recover faithful face images without masks from acquired low-quality face images. Compared with the previous methods which consider image denoising and super-resolution separately, our JDSR-GAN integrates these two tasks together, thus providing collaborative and complementary information to each part, further obtaining pleasing super-resolution results on the benchmark datasets. Comprehensive experimental comparisons have significantly exhibited the superiority of our JDSR-GAN over some approaches in terms of qualitative and quantitative evaluations. 


\bibliographystyle{IEEEtran}
\bibliography{sample-base}

%
\begin{IEEEbiography}[{\includegraphics[width=1in,height=1.25in,clip,keepaspectratio]{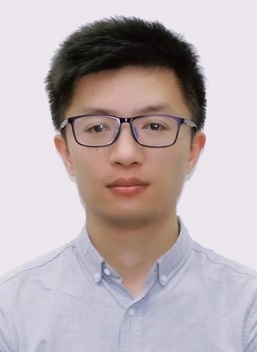}}]{Guangwei Gao}
 (Senior Member, IEEE) received the Ph.D. degree in pattern recognition and intelligence systems from the Nanjing University of Science and Technology, Nanjing, in 2014. He was also a Project Researcher with the National Institute of Informatics, Japan, in 2019. He is currently an Associate Professor in Nanjing University of Posts and Telecommunications. His research interests include pattern recognition and computer vision. He has published more than 60 scientific papers in IEEE TIP/TCSVT/TITS/TMM/TIFS, ACM TOIT/TOMM, AAAI, IJCAI, PR, etc. Personal website: \textit{https://guangweigao.github.io}.
\end{IEEEbiography}

\begin{IEEEbiography}[{\includegraphics[width=1in,height=1.25in,clip,keepaspectratio]{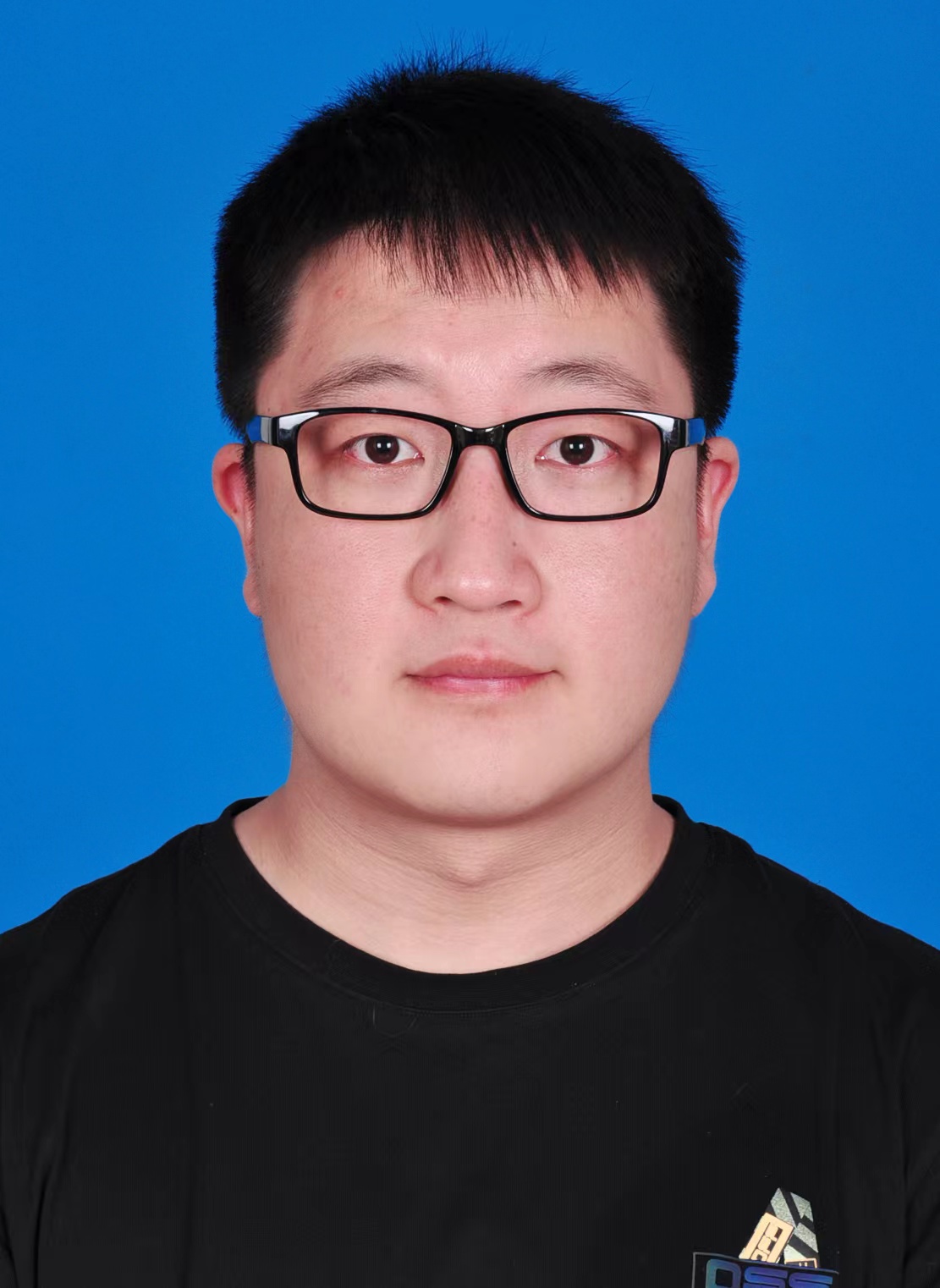}}]{Lei Tang}
 received the B.S degrees in Automation Sciences from Changzhou Institute of Technology, Jiangsu, China, in 2019. He is currently pursuing the M.S. degree with the College of Automation \& College of Artificial Intelligence, Nanjing University of Posts and Telecommunications. His research interests heterogeneous image analysis.
\end{IEEEbiography}

\begin{IEEEbiography}[{\includegraphics[width=1in,height=1.25in,clip,keepaspectratio]{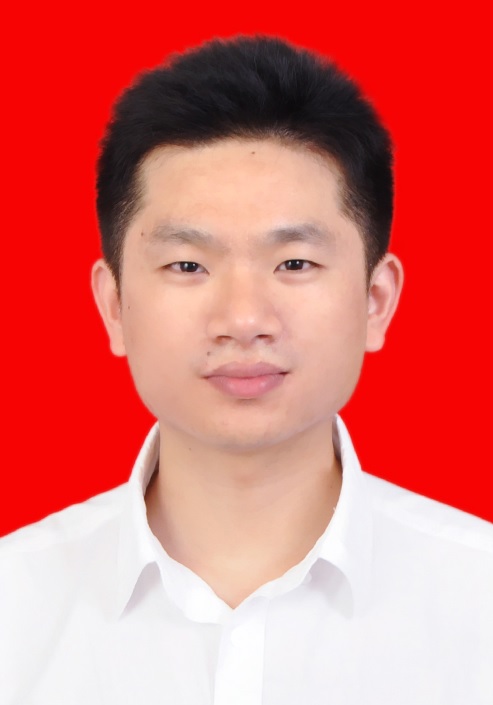}}]{Fei Wu}
 received the Ph.D. degree in Information and Communication Engineering from Nanjing University of Posts and Telecommunications (NJUPT), China, in 2016. He is currently an associate professor with the College of Automation in NJUPT. He has authored over fifty scientific papers. His research interests include pattern recognition and computer vision.
\end{IEEEbiography}

\begin{IEEEbiography}[{\includegraphics[width=1in,height=1.25in,clip,keepaspectratio]{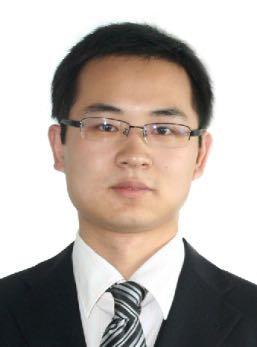}}]{Huimin Lu}
(Senior Member, IEEE) received Ph.D. degree in electrical engineering from the Kyushu Institute of Technology in 2014. From 2013 to 2016, he was a JSPS Research Fellow (DC2, PD, and FPD) with the Kyushu Institute of Technology. He is currently an Assistant Professor with the Kyushu Institute of Technology and an Excellent Young Researcher of MEXT-Japan. His research interests include computer vision, robotics, artificial intelligence, and ocean observing.
\end{IEEEbiography}

\begin{IEEEbiography}[{\includegraphics[width=1in,height=1.25in,clip,keepaspectratio]{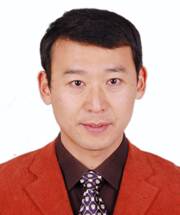}}]{Jian Yang}
(Member, IEEE) received the PhD degree from Nanjing University of Science and Technology (NUST), on the subject of pattern recognition and intelligence systems in 2002. In 2003, he was a postdoctoral researcher at the University of Zaragoza. From 2004 to 2006, he was a Postdoctoral Fellow at Biometrics Centre of Hong Kong Polytechnic University. From 2006 to 2007, he was a Postdoctoral Fellow at Department of Computer Science of New Jersey Institute of Technology. Now, he is a Chang-Jiang professor in the School of Computer Science and Engineering of NUST. His research interests include pattern recognition, computer vision and machine learning. Currently, he is/was an Associate Editor of Pattern Recognition Letters, IEEE Trans. Neural Networks and Learning Systems, and Neurocomputing. He is a Fellow of IAPR.
\end{IEEEbiography}





\end{document}